\documentclass[journal,twocolumn]{IEEEtran}

\usepackage{cite}                            
\usepackage{amsmath,amssymb,amsfonts}
\usepackage{graphicx}
\usepackage{algorithm}
\usepackage{algpseudocode}
\usepackage{xcolor}
\usepackage{booktabs}
\usepackage{makecell}
\usepackage{hyperref}

\makeatletter
\let\@journalNameLogo\relax
\let\@logoname\relax
\makeatother

\usepackage{hyperref}
\usepackage{textcomp}
\usepackage{csquotes}
\usepackage{amsmath}
\usepackage{algpseudocode}
\usepackage{graphicx}
\usepackage{xcolor, soul}
\usepackage{colortbl}
\usepackage{booktabs} 
\definecolor{verylightgreen}{RGB}{220,255,220}
\definecolor{lightblue}{rgb}{0.68, 0.85, 0.9}
\usepackage{makecell}
\usepackage{hyperref}
\usepackage{textcomp}
\def\BibTeX{{\rm B\kern-.05em{\sc i\kern-.025em b}\kern-.08em
    T\kern-.1667em\lower.7ex\hbox{E}\kern-.125emX}}
\def\BibTeX{{\rm B\kern-.05em{\sc i\kern-.025em b}\kern-.08em
    T\kern-.1667em\lower.7ex\hbox{E}\kern-.125emX}}
\begin{document}
\title{Enhancing Early Alzheimer’s Disease Detection through Big Data and Ensemble Few-Shot Learning}
\author{Safa Ben Atitallah, Maha Driss, Wadii Boulila, Anis Koubaa
\thanks{Corresponding Author: Wadii Boulila, (Email: wboulila@psu.edu.sa)}
\thanks{S. Ben Atitallah, M. Driss, W. Boulila, and A. Koubaa are with Robotics and Internet of Things Laboratory, Prince Sultan University, Riyadh 12435, Saudi Arabia.}
\thanks{S. Ben Atitallah, M. Driss, and W. Boulila are with RIADI Laboratory, National School of Computer Science, University of Manouba, Manouba 2010, Tunisia.}
}
\maketitle
\begin{abstract}
Alzheimer’s disease is a severe brain disorder that causes harm in various brain areas and leads to memory damage. The limited availability of labeled medical data poses a significant challenge for accurate Alzheimer’s disease detection. There is a critical need for effective methods to improve the accuracy of Alzheimer’s disease detection, considering the scarcity of labeled data, the complexity of the disease, and the constraints related to data privacy. To address this challenge, our study leverages the power of big data in the form of pre-trained Convolutional Neural Networks (CNNs) within the framework of Few-Shot Learning (FSL) and ensemble learning. We propose an ensemble approach based on a Prototypical Network (ProtoNet), a powerful method in FSL, integrating various pre-trained CNNs as encoders. 
This integration enhances the richness of features extracted from medical images. Our approach also includes a combination of class-aware loss and entropy loss to ensure a more precise classification of Alzheimer’s disease progression levels. The effectiveness of our method was evaluated using two datasets, the Kaggle Alzheimer dataset, and the ADNI dataset, achieving an accuracy of 99.72\% and 99.86\%, respectively.
The comparison of our results with relevant state-of-the-art studies demonstrated that our approach achieved superior accuracy and highlighted its validity and potential for real-world applications in early Alzheimer’s disease detection. 
 
\end{abstract}
\section{Introduction}
\label{sec:introduction}
\IEEEPARstart{A}{lzheimer’s} disease is a progressive neurodegenerative disorder that mainly affects the elderly and causes memory loss and severe cognitive decline.
The advances in medical imaging technologies, such as Magnetic Resonance Imaging (MRI) and Positron Emission Tomography (PET), have opened new avenues for the analysis and understanding of this severe disease \cite{turrisi2023overview,bappi2024novel}. Employing data analytics on these images helps to provide detailed insights about the structural and functional changes in the brain caused by this disease, which facilitates the early diagnosis and monitoring of disease progression \cite{shehab2022machine}.
However, the application of traditional Machine Learning (ML) techniques in analyzing medical images for Alzheimer’s disease diagnosis faces significant challenges \cite{varoquaux2022machine}. One of the primary limitations is the scarcity of labeled data. Medical imaging datasets are usually limited in size because of the high costs associated with data collection, the need for expert annotation, and privacy concerns. This scarcity of labeled data restricts the ability of traditional ML algorithms to learn effectively, as they typically require large volumes of data to achieve high accuracy and generalizability \cite{zhao2023conventional}.
These limitations underscore the need for advanced analytical techniques to leverage the available data more efficiently and extract meaningful patterns from medical images, even in the context of limited labeled datasets.

Few-Shot Learning (FSL) has recently emerged as a breakthrough technology in IoT environments, offering notable benefits, particularly in healthcare and medical research \cite{kotia2021few}. Given the scarcity of labeled data on many medical topics, FSL holds excellent promise. It can learn and make accurate predictions from a few number of samples.  It helps to provide smart solutions with rapid adaptation to new diseases. A trained model on a health use case can quickly adapt to another with minimal data, enabling broad investigations. Moreover, it tackles the big challenge in healthcare research related to privacy concerns by minimizing the data needed for development \cite{wang2020generalizing}. 

In parallel, ensemble learning has emerged as a powerful approach to enhance model performance by combining the predictions of multiple models. Techniques such as voting, boosting, and stacking allow for the aggregation of diverse models to reduce variance and bias and improve generalization \cite{yang2023survey}. Recent advancements in ensemble methods have shown their effectiveness in complex tasks, including image classification and disease diagnosis \cite{mahajan2023ensemble}. By leveraging the strengths of multiple models, ensemble learning can achieve higher accuracy and robustness compared to individual models.

This paper proposes a novel Alzheimer’s disease detection approach that integrates recent advancements in FSL and ensemble learning. Our method employs an ensemble of Prototypical Networks (ProtoNets) with pre-trained Convolutional Neural Network (CNN) encoders to enhance feature extraction and classification performance. Combining class-aware loss and entropy loss ensures precise classification of Alzheimer’s disease progression levels. We evaluate the effectiveness of our method using two datasets, the Kaggle Alzheimer dataset, and the ADNI dataset, demonstrating its superior accuracy and potential for real-world applications.

The primary contributions outlined in this paper are summarized in the following points:
\begin{itemize}
    \item Develop an Alzheimer’s disease detection and progression classification approach using an ensemble of enhanced prototypical networks.
    \item Use pre-trained CNNs, learned on extensive big datasets, to extract image features and enhance the model's ability to discern subtle patterns. 
    \item Integrate a combination of class-aware loss and entropy loss to refine the learning process.
    \item Develop an ensemble of enhanced prototypical networks based on several Transfer Learning (TL) CNN backbones, improving overall performance.
    \item Evaluate the proposed approach using two datasets and compare the results with other DL models and existing studies in the literature to demonstrate the effectiveness of the proposed approach.
\end{itemize}

The subsequent sections of the paper are structured as follows. Section 2 presents a review of related work. Section 3 introduces key concepts relevant to this study, including metric-based learning, TL, and ensemble learning. Section 4 presents in detail the proposed approach. Section 5 describes the used datasets and presents the implementation details and a deep experimental analysis. Finally, Section 6 summarizes the obtained results and outlines potential directions for future research.

\section{Related Work}
Recently, different studies have explored the application of FSL in healthcare, demonstrating the potential of this approach across different medical domains \cite{kotia2021few}.
For the COVID-19 pandemic, the FSL technique has been employed to address the urgent healthcare challenges under conditions of uncertainty and rapid change \cite{lu2021novel}. This approach helped improve the image analysis of X-rays and CT scans and develop efficient diagnostic models based on limited data \cite{jiang2021few,chen2021momentum}. 
Studies in \cite{ouyang2022self,xiao2023boosting,krenzer2023automated,qaraqe2024novel} represent significant advancements in applying ML, particularly FSL techniques, to address specific challenges within the medical field, ranging from medical image segmentation to disease classification and diabetes management. They demonstrate the potential of FSL in enhancing the accuracy and efficiency of medical diagnoses and treatments. 

In parallel, multiple research efforts have explored advanced techniques for detecting Alzheimer’s disease and categorizing its progression stages \cite{mahmood2024alzheimer}. 
The authors in \cite{sharma2022deep} have looked to MRI imaging to diagnose early Alzheimer’s disease. This work used two MRI datasets and a DL algorithm with a VGG16 feature extractor to diagnose Alzheimer. 
Noh et al. in \cite{noh2023classification} introduced a new model for classifying Alzheimer’s disease progression using 4D fMRI data. A U-Net architecture was adopted for spatial feature extraction, and a Long Short-Term Memory (LSTM) network was implemented for temporal feature analysis. Three models with different time-axis inputs were evaluated: 140, 70, and 35 channels, achieving classification accuracies of 96.43\%, 95.71\%, and 91.43\%, respectively. The research demonstrates the potential of combining spatial and temporal features from fMRI data for Alzheimer's classification. 
In \cite{george2024machine}, George et al. examined the effectiveness of several ML models in predicting Alzheimer’s disease. The study investigated reliable prediction models using Kaggle datasets and ML techniques such as Support Vector Machine (SVM), Random Forest (RF), and Gradient Boosting (XGBoost). Two feature extraction approaches were tested: Local Binary Patterns (LBP) and Discrete Wavelet Transform (DWT). Among the models examined, the XGBOOST model with DWT features performed most effectively, with an accuracy percentage of 97.88\%.

Different works have been proposed based on TL models. In \cite{mamun2022deep}, CNN, ResNet101, DenseNet121, and VGG16 were developed for Alzheimer’s disease detection using a dataset of 6219 MRI scans. 
Shukla et al. \cite{shukla2023alz} focused on Alzheimer’s disease classification using CNN models and testing data from the Kaggle repository. Various CNN models were used to classify Alzheimer’s disease phases. For multiclass classification, Alz-MobileConvNet achieved the accuracy of 94\% while Alz-VGGConvNet achieved the accuracy of 99\% for binary classification.
Mujahid et al. \cite{mujahid2023efficient} introduced an efficient ensemble DL model utilizing Adaptive Synthetic (ADASYN). This model integrated VGG16 and EfficientNet, demonstrating significantly high accuracy and AUC scores for multiclass and binary-class datasets.

Based on self-supervised learning, some works have been developed to address challenges related to the limited amount of labeled data or even unlabeled data. Khatri et al. \cite{khatri2023explainable} proposed an explainable vision transformer with SSL to detect the progress of Alzheimer's disease. In \cite{kwak2023self}, Kwak et al. present the Semi Momentum Contrast (SMoCo) framework, a self-supervised contrastive learning approach for predicting the progression of Alzheimer's disease. The method leverages both labeled and unlabeled data to learn general and class-specific representations. Hajamohideen et al in \cite{hajamohideen2023four} introduced a Siamese Convolutional Neural Network (SCNN). SCNN employed a triplet-loss function for the four-way classification of Alzheimer's disease. Both pre-trained and non-pre-trained CNNs are used to generate embeddings for image classification.

As described, researchers have investigated several strategies for recognizing Alzheimer’s disease and identifying its progression phases. DL techniques have shown promise in early Alzheimer’s disease diagnosis, but they frequently struggle with data scarcity and fail to identify changes in brain networks, especially in mild dementia patients. From our review of these related works, we can identify the following main shortcomings. Firstly, the challenges of imbalanced datasets and the scarcity of labeled samples are addressed through simple techniques such as sampling methods. However, sampling can introduce certain limitations, including the potential for over-fitting. While TL has been used in some studies to improve classification accuracy, its effectiveness may be limited, especially when working with very few data samples. Pre-trained models may not fully capture the nuances of Alzheimer’s disease progression across different patient populations. Therefore, there is a need for more robust learning strategies for healthcare datasets to ensure optimal performance. In this work, we investigate using ensemble FSL with pre-trained encoders on big data to handle such challenges.

\section{Background}
This section covers the main major topics involved in our work: metric-based learning, TL, and ensemble learning. 
We use prototypical networks to handle few labeled examples efficiently. By leveraging pre-trained models, we extract valuable representative features. Additionally, we apply ensemble learning to combine various networks, boosting detection accuracy significantly.
\subsection{Metric-Based Learning: Prototypical Network}
Learning from limited labeled data has become more efficient through metric-based learning. Metric-based learning is a flexible and adaptable ML method based on creating an appropriate metric space.
It starts by defining a distance measure between data points. This method allows to estimate the similarities or dissimilarities between data instances by identifying their inherent connections in the metric space \cite{wang2020generalizing}.  
Metric-based learning has a strong ability to adapt to different datasets without the need for extensive retraining. This adaptability makes it particularly well-suited for healthcare scenarios where the data landscape is complex and heterogeneous \cite{he2023few,nayem2023few}.
Different methods based on this type of learning have been proposed and have shown a great enhancement in performance compared to traditional approaches, including Siamese networks, Triplet networks, and ProtoNets.
In this work, we have used the ProtoNet. 

The ProtoNet has been proposed to solve the challenge of FSL, focusing on the overfitting problem related to the limited available data \cite{snell2017prototypical}. 
It is designed to learn a metric space in which data points from the same class are close to each other and separated from data points belonging to different classes. ProtoNets typically comprise a neural network backbone followed by a pooling layer to compute class prototypes. These prototypes represent the central points of each class in the learned metric space. 
Equation (\ref{eq:prototype}) defines how prototypes are computed.
\begin{equation}
    Prototype_{i} = \frac{1}{K} \sum_{j=1}^K f_{\theta_{1}}(x^s_{ij})
    \label{eq:prototype}
\end{equation}
where $x^s_{ij}$ is the samples embedding of the support set, while $k$ specifies the overall number of instances within each class.

Following that, the query samples are categorized by evaluating their distance to the prototypes and attributing them to the class of the closest prototype \cite{xiao2021adaptive}.  
According to Equation (\ref{eq:22}), the output probabilities are determined by applying a softmax function to the negative of the calculated distances.
\begin{equation}
    P(y=c|x) = softmax(-dist(f_{\theta_{1}}(x),Prototype_{i})) 
    \label{eq:22}
\end{equation}
As indicated in Equation (\ref{eq:33}), the loss function $L$ is computed by taking the negative of the natural logarithm of the probability associated with the correct class.
\begin{equation}
    L = - log P_{\theta_{1}} (y = c|x)
    \label{eq:33}
\end{equation}

\subsection{Transfer Learning}
Training DL models with MRI images could be challenging because of their high dimensionality and small sample size \cite{kiziloluk2024eo}. TL has been presented as a solution for this challenge by transferring knowledge from one domain to another \cite{iman2023review}.
Using TL, features from general contexts are reused to serve in more specialized medical domains \cite{atasever2023comprehensive}. Models such as VGG16, ResNet, MobileNet, and EfficientNet have proven to be highly adaptable for such tasks \cite{kim2022transfer}. These models are trained with big data (e.g., ImageNet dataset) and can capture complex patterns and features. 
By applying big data principles, the models accumulated a broad intelligence from millions of generic images. When fine-tuned with medical images, they are used with their pre-learned features which aid in solving the problem of training with small sample sizes.

VGG16 was developed by Oxford University's Visual Geometry Group \cite{simonyan2014very}. This model is known for its depth. It comprises 16 stacked convolutional layers with 3x3 filters, followed by max-pooling layers. This model demonstrates good performance when fine-tuned to recognize intricate patterns in medical imagery, making it highly effective across a wide range of tasks.

The ResNet architecture, proposed by He et al. in \cite{he2016deep}, solves the challenge of vanishing gradients and enables the training of deep networks ideal for complex image recognition tasks. By introducing the concept of residual learning, ResNet facilitates the training of deep networks using skip connections that bypass layers. ResNet has been adapted into different versions, such as ResNet18 and ResNet34, each offering unique advantages while maintaining the core principles of residual learning.

MobileNet \cite{sandler2018mobilenetv2} is specifically designed for mobile and edge devices with limited computational resources. It uses depth-wise separable convolutions, which split the standard convolution into depth-wise and point-wise convolutions, to reduce computational costs while maintaining performance. This architecture is lightweight and efficient for real-time applications on resource-constrained devices.

EfficientNet, proposed by Tan et al. in 2019 \cite{tan2019efficientnet}, provides a scalable architecture that adjusts the depth, width, and resolution of the network. This adaptive approach allows EfficientNet to efficiently balance model complexity and computational resources, resulting in superior performance across various tasks, including medical imaging.

Generally, these models provide the advantage of the powerful DL feature extraction capabilities to the medical imaging domain, where data limitations usually restrict the development of significant ML applications.
Recent studies have investigated the use of TL in medical imaging by fine-tuning pre-trained parameters from non-medical domains \cite{ben2022fusion, ben2022randomly}. 
 
\subsection{Ensemble Learning}
Ensemble learning is a powerful approach for merging the outputs of different DL models and fixing errors presented in individual models, especially in the medical imaging domain \cite{zheng2023application, varone2024finger}.
Ensemble learning is a valuable resource for collective intelligence in ML. Using ensemble learning, we integrate predictions from several models rather than depending on one model. Together, these base learners form a strong ensemble with enhanced performance. 
Different methods have been proposed for ensemble learning, among which Hard Voting (HV) and Soft Voting (SV) are the most prominent techniques for aggregating outputs from other networks.  

\subsubsection{Majority Voting: Hard Decision Method}
The core concept behind the majority voting technique involves choosing the class with the most votes as the optimal output. Predictions generated by various models are gathered and stored in a vector [{$\mathcal P_{1}(x)$, $\mathcal P_{2}(x)$, ..., $\mathcal P_{n} (x)$}], where n represents the number of models. Subsequently, the voting process is employed to determine the output class $y$ for a specific test image. This is achieved by selecting the class most frequently predicted in the vector, as depicted in Equation (\ref{eq:2}):
\begin{equation}\label{eq:2}
    \mathcal Y=  mode {[ P_{1}(x),  P_{2}(x), ... ,  P_{k}(x)]}
\end{equation}

\subsubsection{Weighted Voting: Soft Decision Method}
Weighted voting determines the final output by considering the predicted probabilities $P$ from all models \cite{mohammed2023comprehensive}. The average probability for each class is computed. Suppose we have the following networks: $\mathcal N={ n_{1}, n_{2}, ..., n_{k} }$ employed for multi-classification. Equation (\ref{eq:3}) outlines the method for calculating the average probability of each class.\\
\begin{equation}\label{eq:3}
    \mathcal P_{mean}(i_{j}|x)= \frac{1}{k} \sum_{z=1}^{k} Pn_{z}(i_{j}|x)
\end{equation}

Afterward, the output class for the sample x is established by employing Equation (\ref{eq:9}), which prioritizes the highest probability.

\begin{equation}\label{eq:9}
    \mathcal Y= argmax[P_{mean}(i_{0}|x), ..., P_{mean}(i_{j}|x)]
\end{equation}

\section{Proposed Approach}
In this section, we provide a detailed exploration of the proposed approach, which consists of a set of steps, each designed to contribute to the overall effectiveness.

Our main objective in this study is to develop an efficient model that addresses the problem of labeled data scarcity for Alzheimer’s disease detection and classification. The proposed approach leverages the power of FSL and ensemble learning to learn from limited data and provides enhanced performance with high adaptability in the medical field.
The ProtoNet \cite{snell2017prototypical} is employed in this work. ProtoNet has proven its efficiency in different scenarios by learning a metric space in which classification can be performed by computing distances to prototype representations of each class \cite{kohler2023few}. This method is highly suitable for medical applications, as it helps to develop an efficient model that can be generalized from very few examples \cite{song2023covid,carloni2022applicability,xiao2023boosting}.

\begin{figure*}[h]
    \centering
    \includegraphics[width=0.9\textwidth]{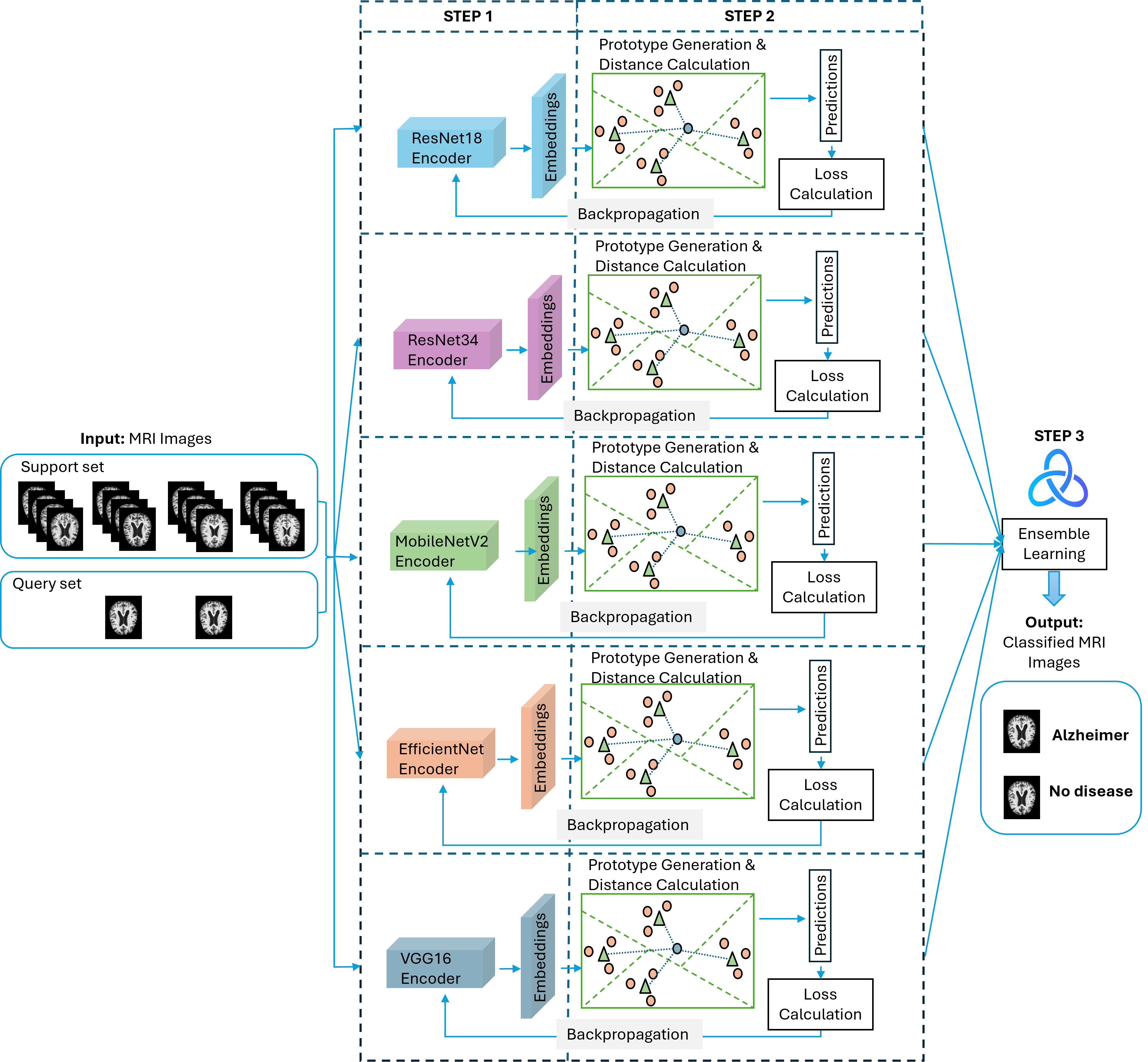}
    \caption{Proposed Ensemble Approach Using Enhanced ProtoNets}
    \label{fig:approach}
\end{figure*}
The proposed approach, as depicted in Fig. \ref{fig:approach}, consists of an ensemble of enhanced ProtoNets. It uses the benefits of pre-trained models for image feature extraction, incorporates the class-aware loss for performance enhancement, and ensemble strategies to increase Alzheimer’s disease detection and classification.
The following will describe each step involved in the proposed approach.

\subsection{Step1: Pre-Trained Models on Big Data}

In the context of ProtoNet, using pre-trained models is presented as a promising approach. 
Pre-trained models, such as VGG16, ResNet, MobileNet, and EfficientNet, have been trained on large-scale datasets and have learned rich hierarchical representations of visual features. Leveraging these pre-trained models as feature extractors in ProtoNets allows them to benefit from the generalization and feature learning capabilities they have acquired during their first training. These pre-trained models serve as encoders, enabling the extraction of highly representative features from MRI images. When applied within ProtoNet, using TL baseline models aids in achieving better performance values compared to the baseline models \cite{singh2021interpretable}. 

In our approach, these pre-trained models on big data serve as a starting point. They are used to capture low-level and high-level features relevant to image recognition tasks. Through TL, we can adapt the previously learned representations to the task of Alzheimer’s disease detection and classification. By fine-tuning the model's parameters on our target dataset, we extract valuable knowledge and, therefore enhance the encoder's capability to represent data comprehensively.

\subsection{Step 2: Enhancing Metric Learning } 
The performance of the ProtoNet model, as a metric-based method, is greatly related to the distribution of instance vectors in the transformed metric space.
However, achieving optimal performance requires addressing the balance between intra-class compactness and inter-class separability within the learned embedding space.
To tackle this challenge, we use Class-Aware Loss (CAL), a loss function for learning discriminative embeddings. CAL is designed to regulate both intra-class compactness and inter-class separability during the training process. Inspired by \cite{liu2023knowledge, xiao2021adaptive}, we employ the farthest positive example and the closest negative example as our optimization targets. In contrast, we enforce a margin between the maximum distance to positive examples and the minimum distance to negative examples for each class prototype. Additionally, CAL introduces a term to ensure that the maximum distance to positive examples does not deviate significantly from the mean distance, promoting a balanced distribution of instance vectors.


First,  the central distance parameter $c$ is computed as the mean of distances from class samples to the class prototype, as follows in equation (\ref{eq:55}):
\begin{equation}
    c = \frac{1}{N_{{pos}}} \sum_{j=1}^{N_{{pos}}} \| \mathbf{x}_{ij} - {Prototype}_i \|_2
\label{eq:55}\end{equation}
where $N_{pos}$ represents the number of positive examples,  and$x_{ij}$ denotes the $j^{th}$ positive example of class $i$.

Next, the maximum distance from the prototype to any positive example ($D_{pro}^{max\_p}$) and the minimum distance from the prototype to any negative example ($D_{pro}^{min\_n}$) are calculated following equations (\ref{eq:66}) and (\ref{eq:77}).
\begin{equation}
    D_{pro}^{max\_p} = \max_{j=1}^{N_{pos}} \| \mathbf{x}_{ij} -Prototype_{i} \|_2
\label{eq:66}\end{equation}

\begin{equation}
    D_{pro}^{min\_n} = \min_{k=1}^{N_{{neg}}} \| \mathbf{x}_{ik} - Prototype_{i} \|_2
\label{eq:77}\end{equation}
  
Based on the obtained constraints, the maximum positive distance and the minimum negative distance, the CAL is calculated using equation (\ref{eq:88}):
\begin{equation}
    L_{ca} = \text{relu}(D_{pro}^{max\_p} -  D_{pro}^{min\_n} + \text{margin}) + 
    \text{relu}(D_{pro}^{max\_p} - c)
\label{eq:88}
\end{equation}

Finally, the CAL is integrated with the cross-entropy loss for cooperative training to discover an enhanced classification boundary. The combined loss is defined as follows in equation (\ref{eq:99}):
\begin{equation}
    L_{comb} = -\log(z_y) + L_{ca}
\label{eq:99}\end{equation}
where $z_y$ denotes the probability that the query instance belongs to class $y$. 

Therefore, by penalizing the maximum distance between features of the same class and their respective prototype, the CAL ensures that features of each class are tightly clustered around their prototype. On the other hand, penalizing the minimum distance between the prototype of a class and the features of other classes ensures that prototypes are well-separated from the negative samples. This helps to get clearer boundaries between different classes.

\subsection{Step 3: Ensemble Learning}
To enhance approach robustness and accuracy, we propose to use an ensemble of ProtoNets. This approach leverages diversity in learning and decision-making by aggregating insights from multiple networks. Each network in the ensemble will focus on different features and aspects of the data, providing a more comprehensive understanding and better generalization to new samples, especially in complex tasks found in healthcare applications.

Five pre-trained big data models, ResNet18, ResNet34, MobileNetV2, VGG16, and EfficientNet, have been selected to get more representative embedding from input data. Each model architecture has unique characteristics and is trained on the extensive ImageNet dataset. They function as feature extractors using their convolutional layers to capture low-level and high-level features from medical imaging data. The extracted features are richer and more generalized, making them significant for subsequent processing stages. 
These features are then fed into a ProtoNet, which is designed to perform well with limited labeled data. This network operates by learning a metric space where classification is performed by computing distances to prototype representations of each class. Furthermore, we employ an ensemble of ProtoNets to enhance the model's performance. Each ProtoNet in the ensemble uses a subset of features extracted by different pre-trained CNNs. This diversity allows the ensemble to capture different aspects of the data, reducing the likelihood of over-fitting and improving generalization to unseen data.
Adopting the concept of ensemble learning, we utilize HV and SV to aggregate the outputs from the five developed ProtoNets. 

In Algorithm \ref{algorithm}, we present the workflow of the proposed approach.
\begin{algorithm}[h]
\caption{\textcolor{black}{Ensemble of enhanced ProtoNets for Alzheimer’s disease detection}}
\label{algorithm}
\begin{algorithmic}[1]
\Require Dataset $D$ with MRI images, $\text{Pretrained\_Models\_List} = [\text{ResNet18,  ResNet34, MobileNetV2, VGG16, EfficientNet}]$
\Ensure Ensemble model for Alzheimer’s disease classification

\State Sample batch:
   $\{(x_i, y_i)\}_{i=1}^N \leftarrow \text{PrototypicalBatchSampler}$

\State Split D into Support and Query Sets:
   \\$S_c = \{(x_i, y_i) \mid y_i = , 1 \leq i \leq K_s\} $\\
   $Q_c = \{(x_i, y_i) \mid y_i = c, K_s < i \leq K_s + K_q\}$
\State Initialize $\text{Prototypes\_List}$ 
\For{each model in Pretrained\_Models\_List}
    \State Instantiate the ProtoNet 
    \State Extract features using the model from $S_c$
    \State Train ProtoNet with extracted features: generate prototypes from $S_c$ using equation\ref{eq:prototype} and calculate distances between $Q_c$ and prototypes using equation\ref{eq:22}
    \State Calculate class-aware loss to refine learning
    \Statex \quad $L_{ca} = \text{relu}(D_{pro}^{max\_p} - D_{pro}^{min\_n} + \text{margin}) + \text{relu}(D_{pro}^{max\_p} - c)$     
    \Statex \quad $L_{\text{comb}} = -\log(z_y) + L_{ca}$   
    \State Add trained model to Prototypes\_List
\EndFor

\State Define Ensemble\_Model that aggregates decisions from Prototypes\_List using voting mechanism
\For{each new sample $x$}
    \State Aggregate predictions from all models in Prototypes\_List
    \State \quad HV: $Y = \text{mode}[P_1(x), P_2(x), \ldots, P_n(x)]$ 
    \State \quad SV: $Y = \arg\max[P_{\text{mean}}(i_0|x), \ldots, P_{\text{mean}}(i_j|x)]$ 
    \State Final\_Decision = Majority\_Vote or Weighted\_Aggregation of predictions
    \State Output Final\_Decision for $x$
\EndFor

\State \Return Ensemble\_Model
\end{algorithmic}
\end{algorithm}




\section{Experiments}
This section describes the used datasets. We also provide and discuss the outcomes of the developed ProtoNets models and the ensemble learning strategies applied to classify Alzheimer's disease. 
\subsection{Datasets}
To validate the proposed approach, the Kaggle Alzheimer 4 classes dataset \cite{Alzheimer_Dataset} was used across the whole experiments. This dataset consists of MRI scans categorized into four classes based on the stage of Alzheimer’s disease progression. 
The dataset's classes are defined below: 
\begin{itemize}
    \item Non-Demented (ND) is the stage of the absence of dementia, where individuals do not show symptoms associated with Alzheimer's disease. The number of samples in this class is 3200.
    \item Very Mild Demented (VMD) is when individuals may have slight memory issues, particularly with recent events. Not all cases of VMD progress to Alzheimer's, but it can be an early sign of the disease. The number of samples in this class is 2240.
    \item Mild Demented (MD) is the stage where  Alzheimer’s disease is usually diagnosed. Symptoms become more noticeable and start to affect daily life. This includes memory loss, changes in personality, difficulty with organizing thoughts, and getting lost. The number of samples in this class is 896.
    \item Moderate Demented (MOD) is characterized by worsening symptoms, including poor judgment, increased confusion, significant memory loss, and needing help with daily activities. The number of samples in this class is 64.
\end{itemize}

To further test the proposed model, we also use the ADNI 3 dataset \cite{ADNI_Dataset} and assess how our model performs across different data. This dataset includes three classes, including mild Alzheimer’s disease Dementia (AD) with 87 samples, which refers to the stage where individuals experience noticeable symptoms of Alzheimer's; Mild Cognitive Impairment (MCI) with 151 samples, characterized by a slight but noticeable decline in cognitive abilities; and Cognitively Normal (CN) with 133 samples, referring to individuals whose cognitive functions are considered normal for their age. Both these datasets are publicly available and have been anonymized to remove any identifiers that could compromise individuals' privacy.



\subsection{Experimental Setup}
We have implemented the proposed approach using a computer with high specifications: an Intel(R) Core(TM) i7-8565U CPU @ 1.80 GHz 1.99 GHz processor, 16 GB RAM running Windows 11, and an NVIDIA GeForce MX graphics card. For coding the DL models, we utilized the Jupyter Notebook from the Anaconda distribution, using Python 3.8. We have also used the PyTorch library, known for being open-source, flexible, and modular. The implementation of the proposed approach is available online at \url{https://github.com/SafaBAtitallah/EnsembleFSL}.

The primary purpose of the suggested approach is to detect and classify the degree of Alzheimer's disease. 
First, we started by preparing data for analytics. We used normalization and resizing for MRI image preprocessing. Each image was normalized to have zero mean and unit variance to reduce the model's sensitivity to image intensity and contrast variations. In addition, images were resized to a standard dimension of 224x224 pixels to ensure uniformity across the dataset and fit the input layer of the pre-trained models. This size provides a balance between computational efficiency and captures sufficient spatial information from images.

Five ProtoNets architectures based on different TL encoders were developed. Each network was trained over 100 epochs to ensure that the models had sufficient iterations to converge to an optimal solution. For faster convergence and better generalization, we configured the networks using the Adam optimizer with a learning rate of 1e-4. We set the embedding output batch size to 128 because larger batch sizes result in quicker convergence and more efficient utilization of computational resources.
To evaluate the proposed model’s performance, a 4-way 10-shot setting is used for the Kaggle Alzheimer dataset. Four test classes are selected for each task, each containing 10 support samples and 15 query samples. For the ADNI dataset, which comprises three classes, a 3-way 10-shot setting was applied, with three test classes selected, each containing 10 support samples and 15 query samples. Table \ref{tab:2} illustrates the hyper-parameters used in our experiments.

\begin{table}[h]
\caption{Employed Hyper-Parameters and Their Values}
\label{tab:2}
\centering
\begin{tabular}{lc}
\toprule
\textbf{Hyperparameter} & \textbf{Value} \\
\midrule
Input size & 224*224*3 \\
Batch size & 32 \\
Embedding size & 128 \\
Epochs & 100 \\
Optimizer & Adam \\
Learning rate & 1e-4 \\
Loss function & Cross-entropy loss + CAL \\
\bottomrule
\end{tabular}
\end{table}

\subsection{Evaluation Performance Metrics}

The effectiveness of the proposed model is evaluated using metrics such as accuracy, precision, recall, and F1-score, as defined in equations (\ref{eq:10} – \ref{eq:13}). 
These metrics are based on the following values:
\begin{itemize}
    \item True Positives (TP): The number of correctly identified positive instances (Alzheimer’s disease).
    \item True Negatives (TN): The number of correctly identified negative instances (non-Alzheimer’s cases).
    \item False Positives (FP): The number of incorrectly identified positive instances.
    \item False Negatives (FN): The number of incorrectly identified negative instances.
\end{itemize}

These metrics thoroughly assess the model's performance, ensuring its reliability and efficiency in Alzheimer’s disease detection and classification.
Additionally, using a confusion matrix offers a visual representation for interpreting the results. 
\\
\textbf{Accuracy:} It evaluates the overall effectiveness of the model across dataset classes.
\begin{equation}\label{eq:10}
    Accuracy=  \frac{TP+TN}{TP+TN+FP+FN}
\end{equation}
\textbf{Precision:} It measures the accuracy of the model in classifying instances as either positive (with Alzheimer's disease) or negative (normal). 
\begin{equation}\label{eq:11}
    Precision=  \frac{TP}{TP+FP}
\end{equation}
\textbf{Recall:} It evaluates the model's ability to correctly detect positive instances.
\begin{equation}\label{eq:12}
    Recall=  \frac{TP}{TP+FN}
\end{equation}
\textbf{F1-score:} It combines the accuracy and recall metrics to provide a value-added measure.
\begin{equation}\label{eq:13}
    F1-score=  \frac{2*Precision*Recall}{Precision+Recall}
\end{equation}
\textbf{Confusion Matrices:} It provides a tabular representation of the actual and predicted classes, allowing for a more detailed analysis of the model's predictions.

\subsection{Experimental Results}

This section explores the various experiment steps, discusses the results, and compares them with previous research.

\begin{figure}[h]
    \centering
    \includegraphics[width = 0.5\textwidth]{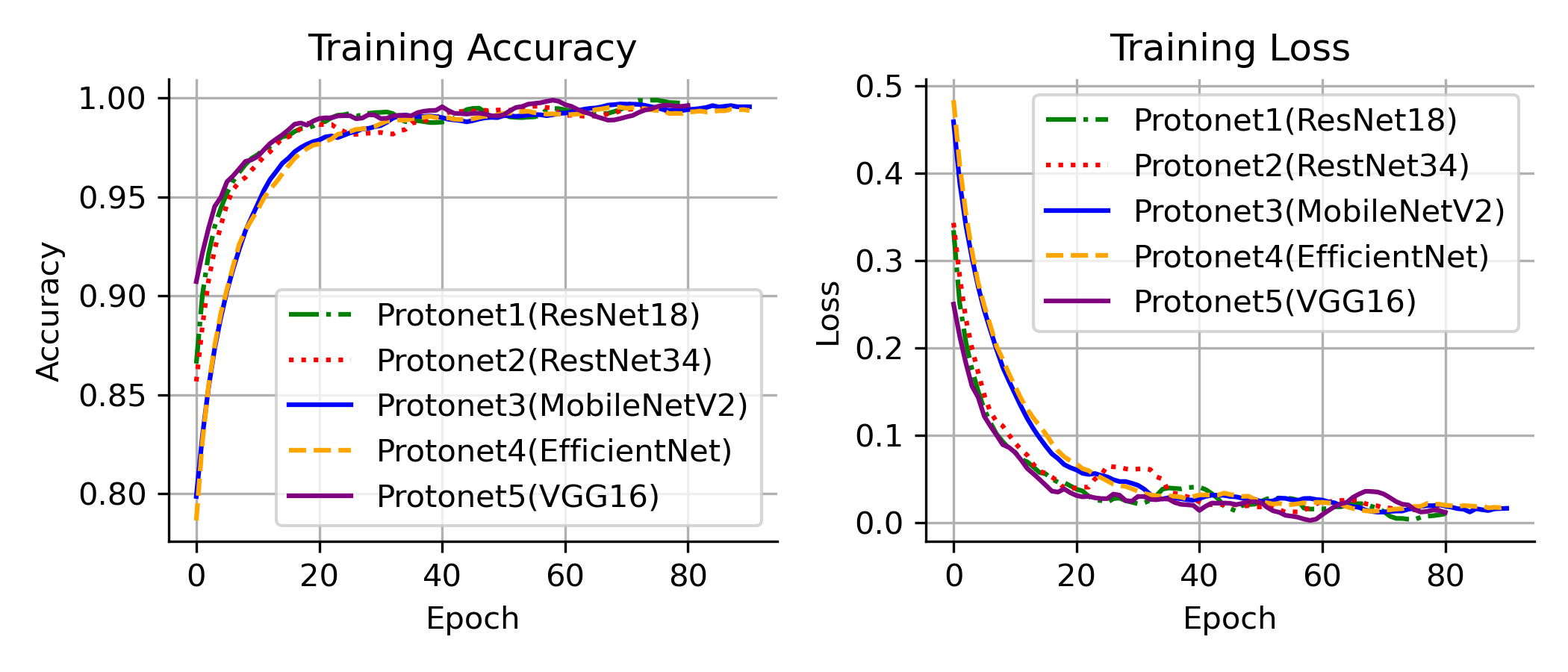}
    \caption{Training Accuracy and Loss of The Developed ProtoNets with Various Pre-trained Models on Kaggle Alzheimer's Dataset. }
    \label{fig:t_plot}
\end{figure}

For both Kaggle and ADNI datasets, we develop five ProtoNets, each of which uses a different pre-trained model on big data as an encoder for feature extraction. 
We train the models without using K-fold cross-validation to avoid issues such as some folds lacking data from certain classes and the increased risk of overfitting.
These models' training accuracy and loss plots with the Kaggle Alzheimer's dataset are shown in Fig. \ref{fig:t_plot}. 
In the left chart, the training accuracy of all models increases over epochs. The curves are closely packed together, showing that each model achieves similar accuracy on the training data. In the right charts, all models show a decrease in training loss over epochs, which indicates that they are learning and improving their predictions of the training data over time. 
After training, the performance of these models is evaluated with unseen test data. 
The experimental results of the trained ProtoNet using the Kaggle dataset are shown in Table \ref{tab:results}. Additionally, we include the results obtained using the ADNI dataset.

The ensemble  ProtoNets model achieved higher performance compared with the other models for both datasets. It attained an accuracy of 99.72\%, and 99.87\% for Kaggle and ADNI datasets, respectively.

Table \ref{tab:results} provides a comprehensive overview of the performance metrics for each developed network. For the first dataset, ProtoNet1, using a ResNet18 encoder, shows a solid performance with an accuracy of 95.51\%. This model also demonstrates a balance across precision, recall, and F1-score.
ProtoNet2, with ResNet34, exhibits slight improvements in accuracy and F1-score over ProtoNet1, indicating that the more complex model can better capture the nuances in the data.
ProtoNet3, utilizing MobileNetV2, and ProtoNet5, with a VGG16 encoder, both show similar performance patterns, indicating that these encoders are quite effective for the Alzheimer detection task but not the best among the tested models.
ProtoNet4, which uses the EfficientNet encoder, stands out with a significant jump in accuracy to 97.57\%. This model tops its counterparts in precision, recall, and F1-score.

 

\begin{table*}[ht]
\caption{\textcolor{black}{Performance Comparison of Enhanced ProtoNets Using Different Encoders Across Two Datasets}}
\label{tab:results}
\centering
\begin{tabular}{@{}lcccccccccc@{}}
\toprule
\textbf{Model} & \textbf{Encoder} & \multicolumn{4}{c}{\textbf{Kaggle Alzheimer Dataset}} & \multicolumn{4}{c}{\textbf{ADNI Dataset}} \\
 & & \textbf{\makecell{Acc. (\%)}} & \textbf{\makecell{Prec. (\%)}} & \textbf{\makecell{Rec. (\%)}}  &\textbf{\makecell{F1 (\%)}} & \textbf{\makecell{Acc. (\%)}} & \textbf{\makecell{Prec. (\%)}} & \textbf{\makecell{Rec. (\%)}}  &\textbf{\makecell{F1. (\%)}} \\
\midrule
ProtoNet1 & ResNet18 & 95.51 & 95.65 & 95.52 & 95.53 & 99.22 & 99.21   &  99.22  & 99.21   \\
ProtoNet2 & ResNet34 & 96.50 & 95.52 & 95.53 & 96.52 & 98.37 &  98.37  &  98.37  &  98.36  \\
ProtoNet3 & MobileNetV2 & 96.11 & 96.24 & 96.11 & 96.12 & 94.58 &  94.68  & 94.58   &  94.60  \\
ProtoNet4 & EfficientNet & 97.57 & 97.59 & 97.58 & 97.58 & 97.39 &  97.41  &  97.39  &   97.38 \\
ProtoNet5 & VGG16 & 96.30 & 96.38 & 96.31 & 96.31 & 97.58 & 97.58   &  97.58  &  97.58  \\
\makecell{Ensemble model} & \makecell{Ensemble of \\ encoders (HV)} & 99.20 & 99.21 & 99.21 & 99.24 & 99.22  &  99.22  & 99.22    &   99.21 \\
\rowcolor{verylightgreen}
\makecell{Proposed model} & \makecell{Ensemble of \\ encoders (SV)} & 99.72 & 99.72 & 99.72 & 99.72 & 99.87 & 99.87     &  99.87  &  99.87  \\
\bottomrule
\end{tabular}
\end{table*}

\begin{figure}[h]
    \centering
    \includegraphics[width=0.27 \textwidth]{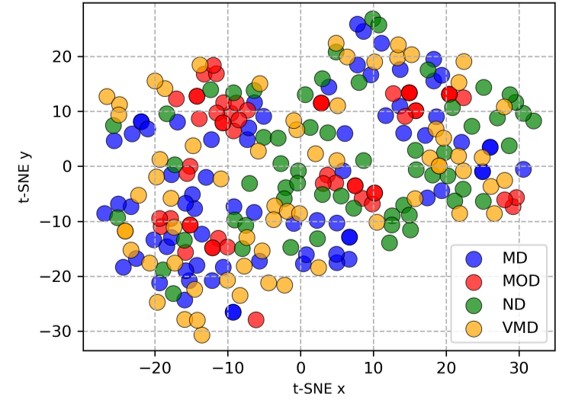}
    \caption{\textcolor{black}{t-SNE Plots Showcasing The Metric Space Representations of The Embedding Before Training}}
    \label{fig:emb_bfr}
\end{figure}

\begin{figure}[h]
    \centering
    \includegraphics[width=0.5 \textwidth]{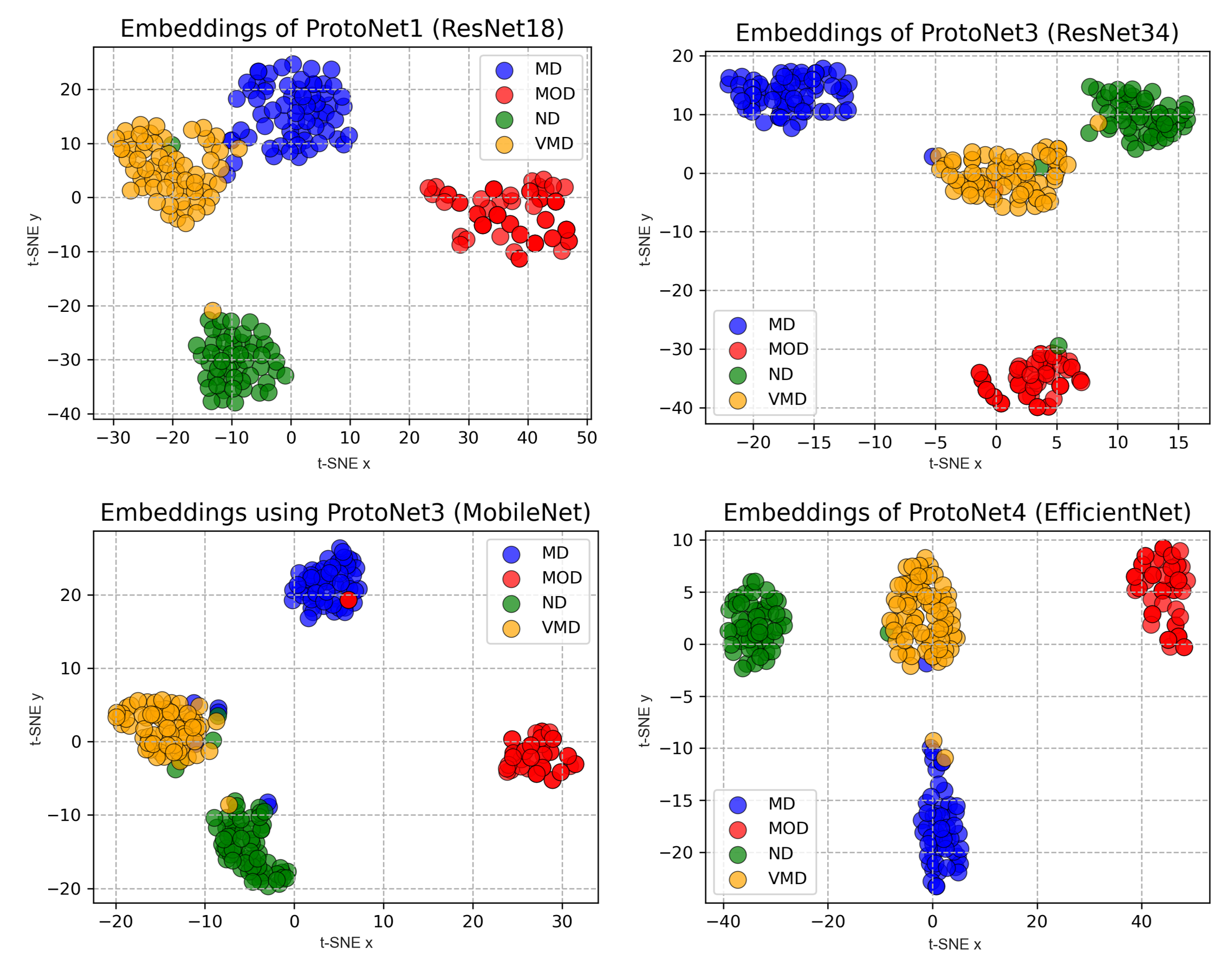}
    \caption{\textcolor{black}{Comparative t-SNE Plots Showcasing The Learned Metric Space Representations For Different ProtoNets}}
    \label{fig:emb}
\end{figure}

To provide a visual presentation of the embedding before training, we include Fig. \ref{fig:emb_bfr}. 
In Fig.\ref{fig:emb}, the t-SNE plots of the trained ProtoNets are depicted, showcasing the learned metric space. We can see the representation of classes in the form of clusters for each of the ProtoNet models. The goal of these plots is to visualize how well each model groups samples from the same class together while keeping samples from different classes separate. We noticed that some samples were misclustered, which will be corrected using ensemble learning.


The Ensemble Model, which combines multiple TL encoders with a HV approach, further increases accuracy to 99.20\%. This performance enhancement indicates that the ensemble method capitalizes on the strengths of individual models and enhances the overall predictive capacity.
The Proposed Model, an ensemble of encoders using SV, achieves the highest accuracy of 99.72\%. This indicates that using an ensemble of encoders and employing a SV mechanism can greatly enhance the model's ability to make accurate predictions. The superior performance of the SV approach can be attributed to its ability to weigh the predictions of each model based on their confidence levels. Unlike HV, which counts the majority votes, SV considers the probability outputs of each model, allowing for a slight aggregation of predictions.
By integrating different feature representations and learning strategies of various pre-trained models, the ensemble approach ensures that the strengths of each ProtoNet encoder are employed resulting in a robust and high-performing model.


\begin{figure}[h]
   \centering
    \includegraphics[width=0.5\textwidth]{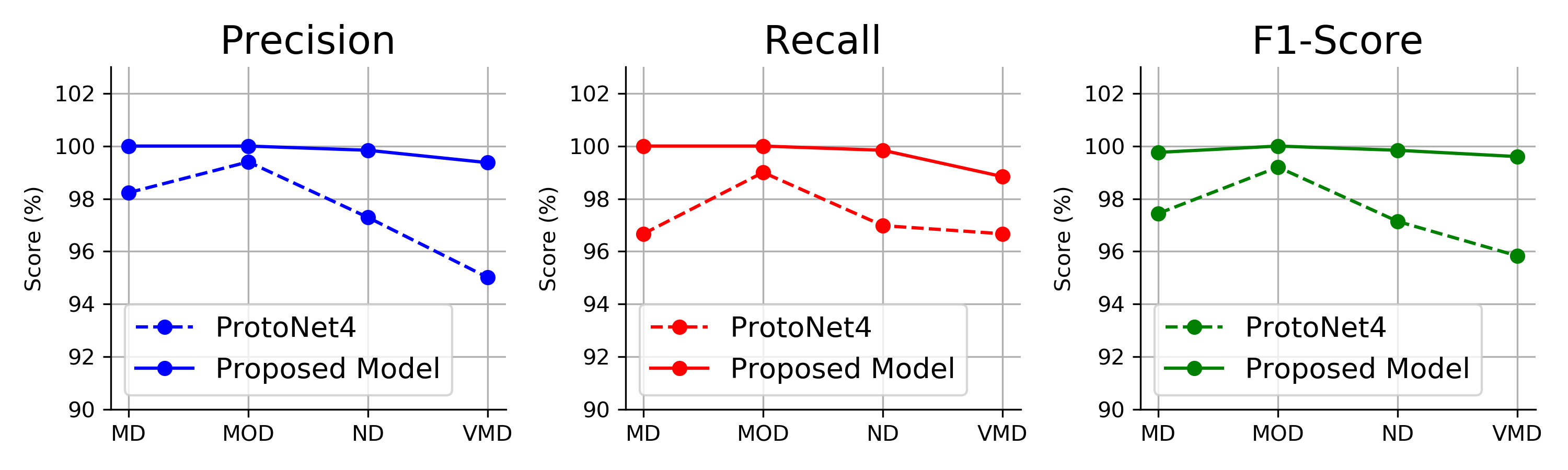}
    \caption{\textcolor{black}{Performance Comparison of The Proposed Model and ProtoNet4 Across the Different Alzheimer's Disease Classes.}}
    \label{fig:comp}
\end{figure}
In Fig. \ref{fig:comp}, we present a comprehensive comparison between the performance of the proposed ensemble model and the ProtoNet variant utilizing the EfficientNet encoder, which has demonstrated the best results among the developed ProtoNets on the Kaggle dataset. The line charts show that the ensemble model, based on SV, exhibits superior performance for precision, recall, and F1-score metrics across the MD, MOD, ND, and VMD categories. 
The performance superiority of the SV ensemble model is attributed to its methodology that integrates multiple classifiers to make a final prediction. Therefore, the proposed model benefits from the collective insights of different encoders, which effectively captures a more diverse representation. Based on these insights, the model could learn an excellent metric space where distances between similar samples are minimized, and distances between different classes are maximized.

\subsection{Ablation Study} 

\begin{figure}[h]
    \centering
    \includegraphics[width=0.5\textwidth]{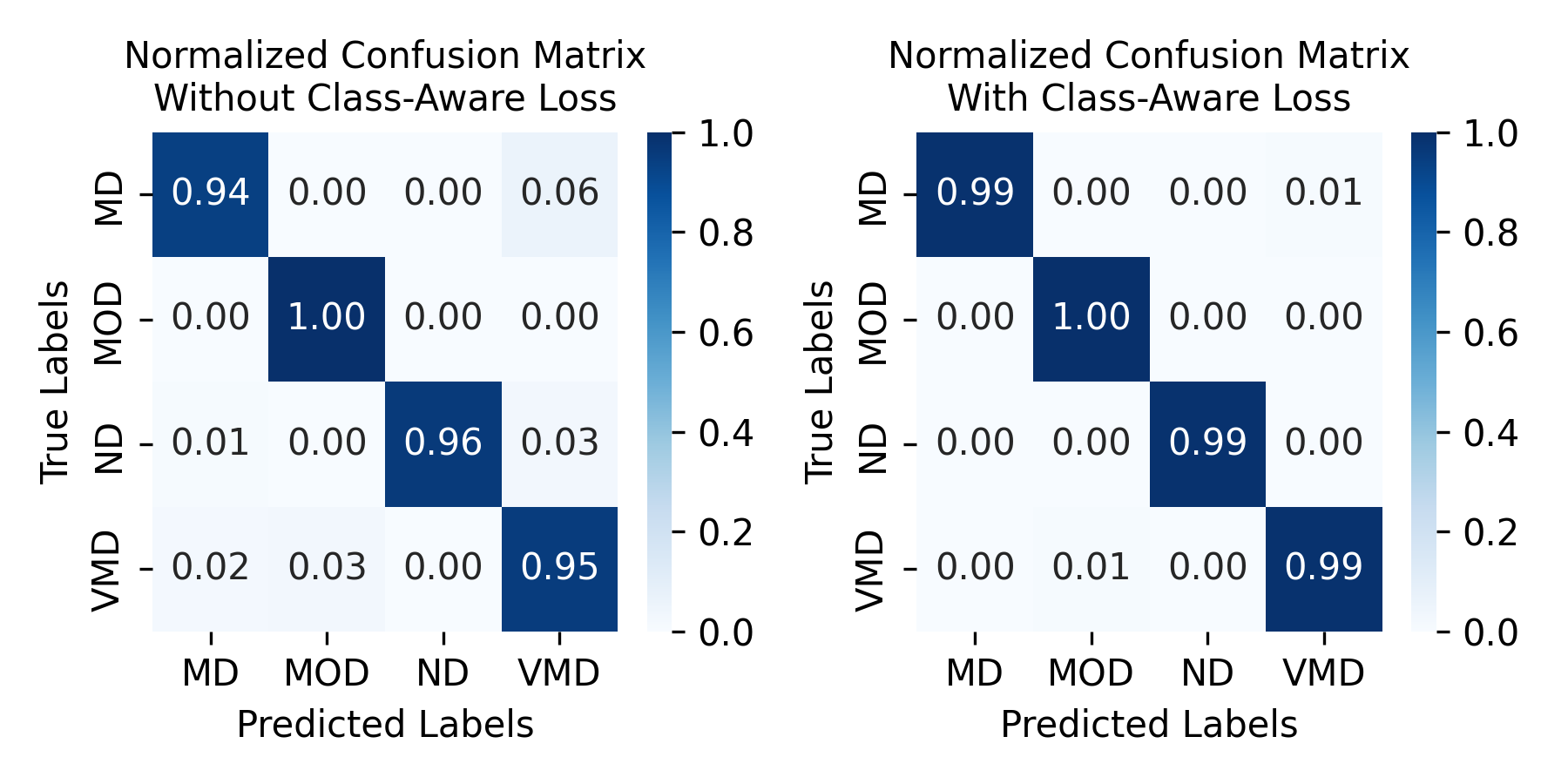}
    \caption{\textcolor{black}{Normalized Confusion Matrices of The Developed Ensemble Models: The SV-based Model, and The Proposed Model. }}
    \label{fig:test}
\end{figure}

To validate the proposed approach, we conduct an ablation study based on the Kaggle dataset, assessing the relevance of the included components.
We examined their impact on model performance by excluding the class-aware loss from the same experiments. 
As illustrated in Table \ref{tab:results2}, the performance of the models has been decreased when it is not used. This demonstrates the role of including 
the class-aware loss function in differentiating between classes and enhancing the learning process. The class-aware loss contributes to obtaining a more discriminative feature space, which also helps achieve higher classification performance. 
The ensemble model based on SV achieved an accuracy of 97.96 \%, with a decrease of about 2\% when compared with the proposed approach.
The confusion matrices of this SV ensemble model and the proposed model are compared in Fig. \ref{fig:test}.
For MOD, both models have performed perfectly, which indicates that features for MOD are easier for the model to learn.
The improvement in the VMD, MD, and ND demonstrates that the used cluster aware loss helps the model to distinguish better between classes that have overlapping features.

\begin{table}[h]
\centering
\caption{Comparison of Performance Metrics Without Class-Aware Loss Across Kaggle Dataset}
\label{tab:results2}
\resizebox{0.48\textwidth}{!}{
\begin{tabular}{@{}lcccccc@{}}
\toprule
\textbf{Model} & \textbf{Encoder} & \textbf{\makecell{Acc. (\%)}} & \textbf{\makecell{Prec. (\%)}} & \textbf{\makecell{Rec. (\%)}}  &\textbf{\makecell{F1 (\%)}}  \\
\midrule
ProtoNet1 & ResNet18 & 93.83 & 93.99 & 93.84 & 93.86      \\
ProtoNet2 & ResNet34 & 93.81 & 93.89 & 93.81 & 93.84    \\
ProtoNet3 & \makecell{MobileNetV2} & 93.31 & 93.43 & 93.35 & 93.12     \\
ProtoNet4 & EfficientNet & 94.39 & 94.21 & 94.39 & 94.53    \\
ProtoNet5 & VGG16 & 92.74 & 92.94 & 92.74 & 92.79      \\
\makecell{HV-based\\ model} & \makecell{Ensemble of\\  encoders} & 96.45 & 96.53 & 96.45 & 96.46    \\
\makecell{SV-based\\ model} & \makecell{Ensemble of\\  encoders} & 97.92 & 98.00 & 97.92 & 97.94   \\
\bottomrule
\end{tabular}}
\end{table}



\subsection{Comparison With Base-Line Models}

Furthermore, Table \ref{tab:performance-results} compares the outcomes with different DL models using the Kaggle dataset. 
Based on the TL approach,  we have fine-tuned ResNet18, ResNet34, MobileNetV2, VGG16, and EfficientNet, and used them for Alzheimer’s disease detection. EfficientNet achieves the highest accuracy and F1-score compared with the other individual pre-trained models, indicating its superior performance in this context.
ResNet18 and ResNet34 achieve relatively similar performance, while MobileNetV2 performs slightly lower than ResNet models and EfficientNet in terms of accuracy and F1-score.
VGG16 shows a lower performance compared to other models.
Besides, we have used ensemble learning to combine these pre-trained models. The ensemble of pre-trained models outperforms individual pre-trained models in terms of accuracy, precision, recall, and F1-score, demonstrating the effectiveness of this method in improving classification performance. Based on the HV method, an accuracy of 78.42\% is achieved. However, employing the SV method yields a higher accuracy of 79.54 \%.
Another comparison was conducted using the conventional ProtoNet, which is based on a simple CNN architecture as a backbone (encoder). It achieved a performance of 89.96\%, demonstrating the capacity of FSL methods to quickly learn from a small number of examples compared to pre-trained models.  

From this comparative analysis, we can highlight the following key findings:
\begin{itemize}
    \item Limitations of TL models: Employing pre-trained models alone does not effectively address the challenges of Alzheimer's detection, particularly due to imbalanced datasets and the limited number of samples in some classes. Pre-trained models fail to generalize well to under-represented classes, leading to low performance in classification tasks.
    \item Advantages of ensemble learning: Ensemble learning emerges as a powerful technique for enhancing performance in Alzheimer's detection. By combining predictions from multiple models, ensemble methods can leverage the strengths of diverse classifiers, leading to improved performance.
    \item Relevance of ProtoNets: The use of ProtoNets emerges as a promising solution for addressing the challenges of Alzheimer's detection. Unlike TL models, ProtoNets offer a more suitable approach for working with imbalanced datasets and a limited number of samples in some classes. 
\end{itemize}




\begin{table}[htbp]
\centering
\caption{\textcolor{black}{Comparison of Performance Metrics for Different DL Models Across Kaggle Dataset}}
\label{tab:performance-results}
\resizebox{0.48\textwidth}{!}{%
\begin{tabular}{@{}ccccccc@{}}
\toprule
\textbf{Model} & \makecell{\textbf{Ensemble}\\ \textbf{Method}} & \textbf{\makecell{Acc. (\%)}} & \textbf{\makecell{Prec. (\%)}} & \textbf{\makecell{Rec. (\%)}}  &\textbf{\makecell{F1. (\%)}} \\
\midrule
ResNet18 & - & 75.37 & 76.21 & 75.45 & 75.12  \\
ResNet34 & - & 72.69 & 72.15 & 72.41 & 71.96  \\
MobileNetV2 & - & 71.54 & 71.15 & 71.54 & 70.80  \\
VGG16 & - & 69.27 & 68.30 & 66.27 & 67.43  \\
EfficientNet & - & 77.87 & 78.45 & 77.87 & 77.71  \\
\makecell{Ensemble \\TL models} & HV & 78.42 & 78.23 & 78.42 & 78.11 \\
\makecell{Ensemble \\TL models} & SV & 79.54 & 79.34 & 79.12 & 79.61 \\
\rowcolor{verylightgreen}
\makecell{ProtoNet \\ (CNN backbone)} & - & 89.96 & 89.14 & 89.95 & 89.73 \\
\bottomrule
\end{tabular}}
\end{table}

\subsection{Comparison With Related Works}
This study used FSL and ensemble learning to create an effective Alzheimer’s disease detection and classification approach. 
The proposed method was evaluated against recent techniques reported in the literature that have used the same datasets for evaluation.
Table \ref{tab:RW_comp} summarizes the comparison of performance results in terms of accuracy, precision, recall, and F1-score. According to Table \ref{tab:RW_comp}, our proposed approach outperforms the state-of-the-art approaches, achieving the highest accuracy of 99.72\%. This indicates the effectiveness of the proposed approach in Alzheimer’s disease detection and demonstrates its superiority over existing techniques documented in the literature that were based mainly on TL.
While TL method has been widely adopted to overcome the scarcity of domain-specific data, it often require fine-tuning with a substantial amount of target domain data to achieve optimal performance.

The superior performance of the proposed approach is attributed to the utilization of FSL and ensemble learning techniques. These advanced methods enable the model to effectively learn from limited data and leverage the complementary strengths of multiple models, resulting in a more robust and accurate diagnostic system. 
Additionally, this performance is also related to several key factors used to enhance the ProtoNet. Firstly, the high representation of features extracted by different pre-trained model encoders plays an important role in capturing the diverse and informative aspects of the input data. Secondly, integrating class-aware loss aids in refining the learning process and encourages the model to learn distinct clusters for each class. Finally, ensemble learning combines the predictions of multiple models, leverages their complementary strengths, and mitigates individual model biases.    


\begin{table}[ht]
\centering
\caption{Performance Comparison of the Proposed Model with Related Works}
\label{tab:RW_comp}
\resizebox{0.49\textwidth}{!}{%
\begin{tabular}{ p{1cm} c p{2.8cm} ccccc }
\toprule
\textbf{Work} & \textbf{Dataset} & \textbf{\begin{tabular}[c]{@{}c@{}}Method\end{tabular}} & \textbf{\begin{tabular}[c]{@{}c@{}}Acc. \\(\%)\end{tabular}} & \textbf{\begin{tabular}[c]{@{}c@{}}Pre.\\ (\%)\end{tabular}} & \textbf{\begin{tabular}[c]{@{}c@{}}Rec. \\(\%)\end{tabular}} & \textbf{\begin{tabular}[c]{@{}c@{}}F1. \\(\%)\end{tabular}} \\ 
\midrule
\begin{tabular}[l]{@{}l@{}}Sharma \\et al. \cite{sharma2022deep} \\(2022)\end{tabular} & Kaggle & NN classifier with a VGG16 feature extractor & 90.4 & 90.5 & 90.4 & 90.4 \\
\begin{tabular}[l]{@{}l@{}}Kwak\\ et al. \cite{kwak2023self}\\ (2023)\end{tabular} & ADNI & Semi Momentum Contrast framework & 81.09 & 82.17 & - & - \\
\begin{tabular}[l]{@{}l@{}}Hajamo-\\hideen \\et al. \cite{hajamohideen2023four}\\ (2023)\end{tabular} & ADNI & Siamese Convolutional Neural Network& 91.83 & - & - & - \\ 
\begin{tabular}[l]{@{}l@{}}Noh \\et al. \cite{noh2023classification} \\ (2023)\end{tabular} & Kaggle & U-Net architecture with LSTM & 96.4 & 96.96 & 96.76 & 96.79 \\
\begin{tabular}[l]{@{}l@{}}George \\et al. \cite{george2024machine}\\(2024)\end{tabular} & Kaggle & Gradient boosting with Discrete Wavelet Transform features & 97.88 & - & 98.13 & 97.91 \\ 
\rowcolor{verylightgreen}
\begin{tabular}[l]{@{}l@{}}Proposed \\ model\end{tabular} & \begin{tabular}[c]{@{}c@{}}Kaggle\\ ADNI\end{tabular} & Pretrained models, FSL, and ensemble learning & \begin{tabular}[c]{@{}c@{}}99.72\\99.87\end{tabular} & \begin{tabular}[c]{@{}c@{}}99.72\\99.87\end{tabular} & \begin{tabular}[c]{@{}c@{}}99.72\\99.87\end{tabular} & \begin{tabular}[c]{@{}c@{}}99.72\\99.87\end{tabular} \\
\bottomrule
\end{tabular}}
\end{table}

\subsection{Complexity Analysis}
The computational complexity of our approach is influenced primarily by the components of our proposed model. Initially, we utilize pre-trained models for feature extraction, which are inherently more complex than simpler neural networks. Despite their computational demands, these models provide significantly richer and more representative feature sets from medical images. In the subsequent phase, our ProtoNets compute distances within a metric space for classification. This process involves fewer parameters compared to conventional ML models and requires distance calculations, which become notably intensive only when dealing with a large number of classes and samples. In the final phase, our approach integrates multiple ProtoNets to form an ensemble. While this increases the computational load linearly with the number of included models, it boosts the robustness and accuracy of the system, thereby justifying the additional computational expenses.

\subsection{Discussion}
Several challenges are associated with the early detection and classification of Alzheimer’s disease progression, including the privacy of patients, the scarcity of labeled samples, the difficulties in collecting extensive datasets, and the high costs associated with such efforts. To face these challenges, this paper presents a novel approach leveraging big data, FSL, specifically ProtoNets, and ensemble learning. The FSL technique allows for learning rich representations from a limited number of samples, achieving high accuracy and facilitating the classification of MRI images to detect abnormalities.
The conventional methods for medical image analysis, heavily reliant on large labeled datasets, are ineffective in the context of Alzheimer's disease, where both accessibility and ethical considerations constrain data collection. ProtoNets excels in classification tasks with little data. This capability is particularly advantageous in medical fields where data scarcity and the need for precise classifications are perennial issues.
Our experimental studies underscore the efficacy of ProtoNets in enhancing the performance of Alzheimer’s disease detection and classification tasks. By integrating pre-trained models on big data for feature extraction, networks achieve highly representative embeddings. Additionally, class-aware loss refines the learning process further and identifies class boundaries within the embedding space. Finally, combining five ProtoNets, each utilizing different encoders, significantly enhances results. Through comprehensive comparative analysis, we showcase the superiority of our approach over existing models. Moreover, our method outperforms related works, demonstrating its higher performance.

Despite these advancements, it is important to recognize the limitations of our approach, including data dependency, computational complexity, and interpretability. The performance of our model depends on the quality of the available datasets, and the variations in MRI images can impact its generalizability. The use of multiple pre-trained CNNs and an ensemble of ProtoNets increases computational complexity and training time, which limits its applicability in resource-constrained environments. Additionally, the proposed model acts as a black box, making decision interpretability challenging. Addressing these limitations in future work is essential for advancing Alzheimer’s disease detection and ensuring the practical applicability of computer-aided methods in clinical settings.
\section{Conclusion}

Alzheimer's disease is a profoundly severe neurological disorder. 
Detecting Alzheimer’s disease in its early stages is indeed for initiating timely interventions and treatments to slow down the progression of the disease and improve the quality of life for affected individuals and their families. 
This paper presents an effective approach for detecting and classifying Alzheimer’s disease progression using an ensemble of enhanced ProtoNets. Our method demonstrates improved accuracy and precision in classifying Alzheimer’s disease progression levels by employing pre-trained CNNs on big data for feature extraction and integrating refined loss functions. By developing an ensemble of networks, we achieve superior performance using the weighting voting method.
Evaluation with the Kaggle Alzheimer and ADNI datasets achieved promising results, with an accuracy of 99.72\% and 99.86 \%.

In response to the growing need for explainable AI, particularly in healthcare settings, our next phase of development will focus on enhancing the interpretability of our model. 
Additionally, we aim to validate the application of our approach in real-time settings, enhancing its capability to detect and respond instantly in clinical settings. Furthermore, we seek to explore the adaptation of our approach to addressing related medical imaging tasks, including the detection and classification of other neurological disorders or medical conditions with limited labeled data.

\section*{Acknowledgments}

The authors would like to thank Prince Sultan University for their support.

%
%
\bibliographystyle{IEEEtran} 
\bibliography{references.bib}  
\end{document}